# Soccer League Optimization: A heuristic Algorithm Inspired by the Football System in European Countries

Erfan Khaji

Graduate Student, Complex Adaptive Systems Group, University of Gothenburg, Gothenburg, Sweden.

**Abstract**

In this paper a new heuristic optimization algorithm has been introduced based on the performance of the major football leagues within each season in EU countries. The algorithm starts with an initial population including three different groups of teams: the wealthiest (strongest), the regular, the poorest (weakest). Each individual of population constitute a football team while each player is an indication of a player in a post. The optimization can hopefully occurs when the competition among the teams in all the leagues is imitated as the strongest teams usually purchase the best players of the regular teams and in turn, regular teams purchase the best players of the weakest who should always discover young players instead of buying professionals. It has been shown that the algorithm can hopefully converge to an acceptable solution solving various benchmarks.

## Introduction

Several heuristic optimization algorithms have been introduced since two decades ago. The heuristic algorithm have been extremely noticed and considered recently for several reasons since the computational costs have been decreased utilizing high performance computation skills. Algorithms such as Genetic Algorithm (GA) [1], Particle Swarm Optimization (PSO) [2], Simulated Annealing (SA) [3], Ant Colony Optimization (ACO) [4,5], Imperialist Competitive Algorithm (ICA)[6], and Bee Colony Optimization [7] are those heuristic methods based on natural or socio-political optimization processes occurring within our surrounding.

GA is in fact, based on the genetic development of human being as better genes in a population are being combined and mutated for the next generations which guarantee the improving of the population fitness within the time. PSO and other swarm intelligence algorithms, on the other hand, seek to imitate the behavior of animals living in a society e.g. fishes swarm, bee swarms, ant colony or birds swarms. The rules which guarantee the group to stay together while each individual has its own sense of guidance are the key points in such algorithms. Another widely used heuristic algorithm, SA, is based on a scientific observation in metallurgy engineering as a substance with higher temperature of its melting point gradually lose its temperature until it create crystalline lattice.

An example of socio-political algorithm in optimization is ICA inspired by the imperialist completion of European powers within the nineteen century. The biggest powers absorb more colonies compared to weaker powers which hopefully converge to an acceptable solution for global optimum within the time.

All the mentioned algorithms have several applications in combinatorial, non-linear, and non-convex optimization problems which are too time-wasting in order to be solved with classical methods, or the nature of which make them impossible to solve.

In this work, the optimization process is inspired by the optimization of football systems in European countries where the best players usually sold to the wealthiest clubs. The poorest clubs are financially limited which make them to discover young players and train them without paying for new players. The last sorts of teams are the ones which have a combination of these two policies for players. They buy good players of the poorest and worse players of the wealthiest teams. The system within the time, improve the whole football system of such country as it is currently occurring in Germany, France, Spain or Italy. The algorithm is introduced in the second section while the computational results and the conclusion are presented in sections 3 and 4.

## The proposed method

A football system in an EU country including several leagues e.g. Italy (Serie A, B, C1,..), Germany (Bundesliga 1, Bundesliga 2…), etc. Each league is consisting several teams in a way that the more wealthier a team in a league, the better player it afford to purchase and the less possibility to discover younger players. The wealthiest teams try to track perfect players in less important teams who have perfect performance in their clubs and hire them. They usually try to improve those players' performance and wait to see their outcome. Should any player show non-satisfactory performance, they will be replaced ASAP.

Regular clubs who are working in lower levels have two options: buying not-perfect players of better clubs or perfect players of worse clubs, or discover younger players. The poorest clubs always are limited to discover young players by help of which they can make money since if they show valuable performance, they will be transferred into better clubs, and otherwise, they should be replaced by other younger players.

The main idea here is the fact that in each football system, there are those clubs who just discover young players and sell them to wealthier clubs (the wealthiest and the regulars). In addition, there are clubs who buy a combination of players from the strongest or the weakest teams or discover their own young players. And finally, there are few perfect clubs who enjoy the most perfect outcomes of the rest clubs and buy them.

Suppose the following comparisons: A cell of number: a player, an array of numbers: a team, adding a range of random numbers to cell: training a player, generating a random number: discovering a young player, and a dimension: a post in a team. Considering the mentioned assumptions, we can conclude the following algorithm in addition to its diagram.

1- Determination of number of teams in each class: Na: number of the wealthiest teams, Nb: number of regular teams, and Nc: number of the weakest teams, and N: number of seasons.
2- Generating a random population for all three levels. (creating teams)
3- Classifying the teams into three levels based on their primary performance $f(x)$. (primitive evaluation)
For season 1: N
a- Determination of one of the dimensions for each number (a player of each team)
b- Adding different random numbers to each selected dimension (training the selected player)
c- Limiting the dimensions to be in selected range of variables (learning relative skills according to that league e.g. defending in Italy or crossing in England)
d- Computing $f'(x)$ (measuring the gathered points at the end of the season after buying the selected player)
e- Saving the best performance of the system's history.
f- Comparing the points of the teams before and after buying the players (calculating $f'(x)-f(x)$).
g- $F(x)=f'(x)$
h- Determination the players who have improve their teams overall points and vice versa.
i- For the wealthiest teams, those who have improved their teams' score will be remained for the next season while those who have shown bad performance will be replaced by outstanding players of regular teams. For regular teams, those who have improved their teams' scores will be sold to the wealthiest while weaker players will be replaced by good players from the weakest. Finally, perfect players of the weakest teams will be sold to regular clubs while the worse one will be replaced by new discovered players.

There are some characteristics about SLO algorithm. Firstly, the exploration ability of the algorithm is notably high based on the fact that the wealthiest teams are greed for outstanding players which may or may not help them to improve their scores. The following figures show the positions of the wealthiest teams and regular teams before and after N seasons on the benchmark G4.

**Figure 1: Positions of the wealthiest teams before starting the seasons on the benchmark G4.**

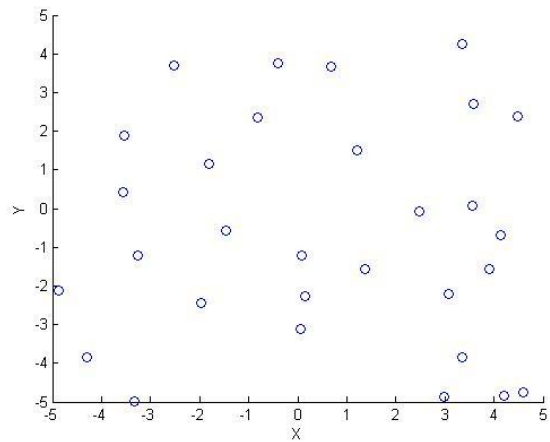

Figure 2: Positions of the wealthiest teams after N seasons on the benchmark G4. The red point is the outcome of the algorithm. Although there are some teams whose players are near the best team's players in terms of quality, the other wealthy teams still have far distance from the best team. This point is near the truth since in reality, not all the wealthiest teams are benefited from their expensive transfers.

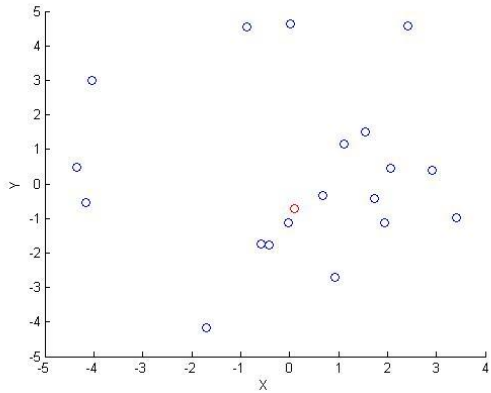

Figure 3: The positions of the swarms before their travel on the surface of benchmark G4.

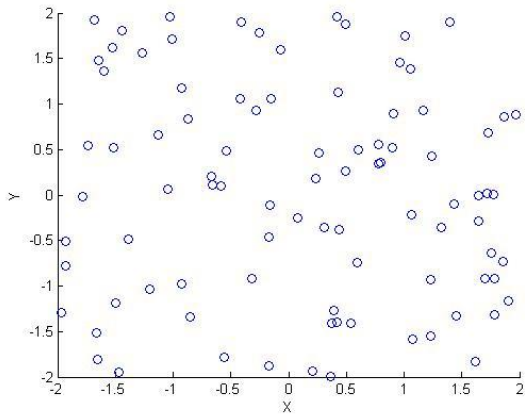

Moreover, the algorithm pays specific attention to each dimension itself. Opposed to other heuristics where the points placed closer to best answers generally, better points in SLO absorbs better dimensions instead of whole the point in contrast with GA (cross over), PSO, and ICA.

Figure 4: The positions of the swarms before their travel on the surface of benchmark G4. The red line is the outcome of applying PSO on the benchmark G4. The figure indicates on the further convergent of all the swarms compared to SLO i.e. less exploitation or randomness ability.

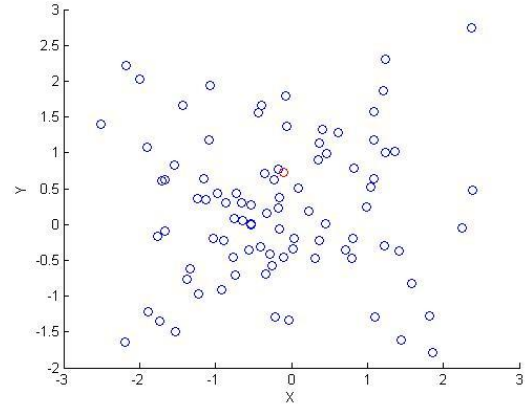

## Computational Results

Four different benchmarks were tested as instances to examine the exploration ability of the algorithm. One of the benchmarks war also tested with PSO and GA. Number of all three sorts of teams were equal to 30 which was similar to 90 swarms in PSO and 90 genes in GA. Number of seasons was considered to be 100. For each example, the benchmark function in addition to its optimum is presented. The only remained point is the fact that GA and SLO are designed to solve the maximization problem i.e. in soccer leagues the best team obtained more points. Therefore, for minimization problems, SLO tends to solve $-f(x)$. Each benchmark solved for 5 times. The optimum points in addition to their positions are presented for each benchmark.

**1:** $G1=(1.5-x*(1-y))^2 + (2.25-x*(1-(y^2)))^2 + (2.625 - x*(1 - (y^3)))^2.$

Table 1: Results of 5 tests on problem G1.

| No of Test | F(Global Minimum) = 0. | Global Minimum= (3.000  0.5) |
|---|---|---|
| 1 | -9.2540e-004 | 3.0763  0.5167 |
| 2 | -2.6928e-004 | 3.0401  0.5108 |
| 3 | -1.0979e-004 | 3.0118  0.5010 |
| 4 | -0.0015, | 2.9111  0.4746 |
| 5 | -2.6301e-004 | 2.9615  0.4895 |

**Figure 5:** The cost of the best poorest teams in each season for the problem of G1 where Na=Nb=Nc=30 and Number of seasons is 100.

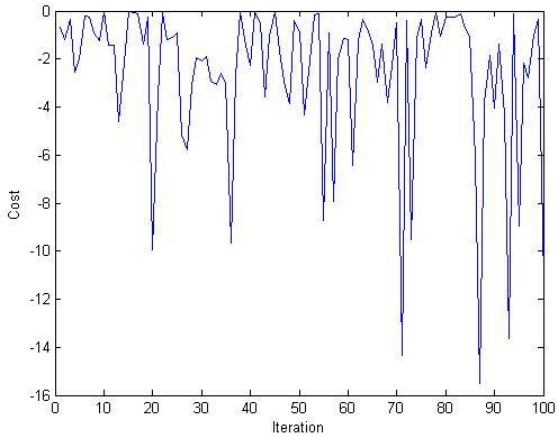

**Figure 6:** The cost of the best regular teams in each season for the problem of G1 where Na=Nb=Nc=30 and Number of seasons is 100.

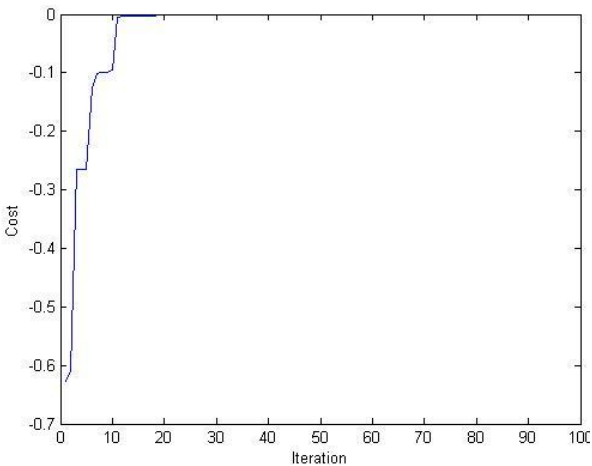

**Figure 7:** The cost of the best wealthiest teams in each season for the problem of G1 where Na=Nb=Nc=30 and Number of seasons is 100.

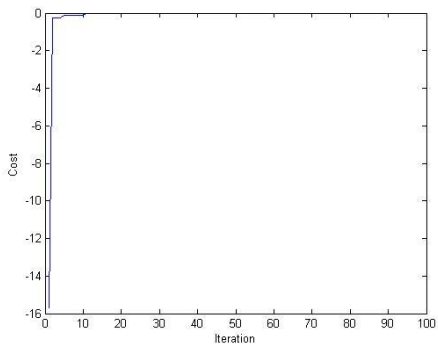

2: G2= (1 + ((x + y + 1)^2)*(19 - (14*x) + (3*(x^2)) - 14*y + 6*x*y + 3*(y^2)))*(30 + ((2*x - 3*y)^2)*(18 - 32*x + 12*(x^2) + 48*y - 36*x*y + 27*(y^2)))

**Table 2:** Results of 5 tests on problem G2.

| No of Test | F(Global Minimum)= -3 | Global Minimum= (0 -1) | |
|---|---|---|---|
| 1 | -3.0001 | 0.0003 | -1.0003 |
| 2 | -3.0002 | -0.0009 | -1.0005 |
| 3 | -3.0017 | 0.0028 | -0.9991 |
| 4 | -3.0041 | 0.0004 | -0.9968 |
| 5 | -3.0001 | 0.0006 | -0.9998 |

**Figure 8:** cost of the best wealthiest teams in each season for the problem of G2 where Na=Nb=Nc=30 and Number of seasons is 100.

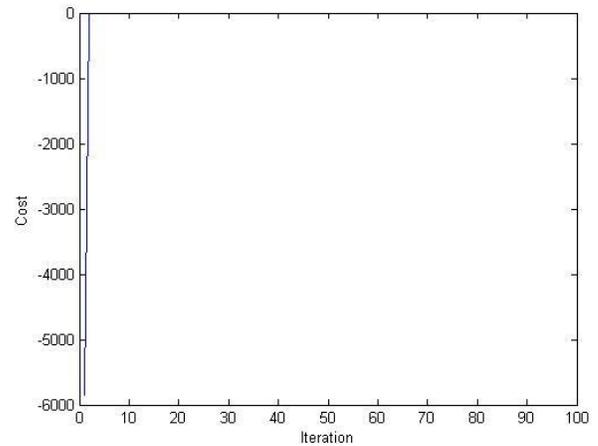

3: G3: 1+(-13+x-(y^3)+(5*(y^2))-(2*y))^2+(-29+x+(y^3)+(y^2)-(14*y))^2

**Table 3:** Results of 5 tests on problem G3.

| No of Test | F(Global Minimum)= -1 | Global Minimum= (5 4) | |
|---|---|---|---|
| 1 | -1.0014 | 4.9824 | 4.0010 |
| 2 | -1.0002 | 5.0122 | 3.9997 |
| 3 | -1.0048 | 4.9723 | 4.0019 |
| 4 | -1.0048 | 4.9730 | 4.0019 |
| 5 | -1.0002 | 5.0075 | 3.9996 |

**Figure 9: cost of the best wealthiest teams in each season for the problem of G3 where Na=Nb=Nc=30 and Number of seasons is 100.**

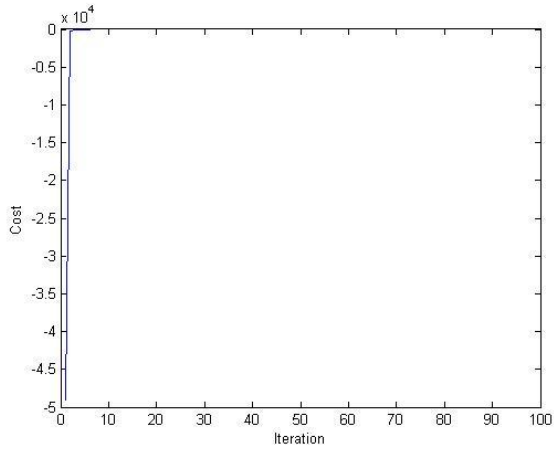

**Figure 10: Cost of the best fitness in each iteration for the problem of G3 where number of swarms=90 and Number of iterations is 100.**

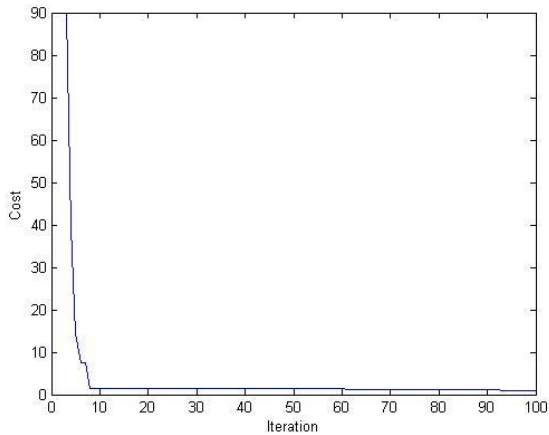

**Figure 11: Cost of the best fitness in each iteration for the problem of G3 where number of genes is 90 and Number of iterations is 100.**

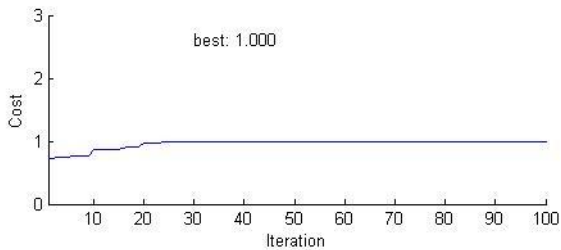

4: $G4 = 4*(x^2) + 2.1*(x^4) + ((x^6)/3) + (x*y) - 4*(y^2) + 4*(y^4)$

**Table 4: Results of 5 tests on problem G4.**

| No of Test | F(Global Minimum)= -1 | Global Minimum= (5 4) |
|---|---|---|
| 1 | 1.0316 | -0.0899   0.7128 |
| 2 | 1.0316 | 0.0886   -0.7127 |
| 3 | 1.0316 | -0.0911   0.7136 |
| 4 | 1.0316 | 0.0888   -0.7120 |
| 5 | 1.0316 | -0.0917   0.7136 |

**Figure 12: cost of the best wealthiest teams in each season for the problem of G4 where Na=Nb=Nc=30 and Number of seasons is 100.**

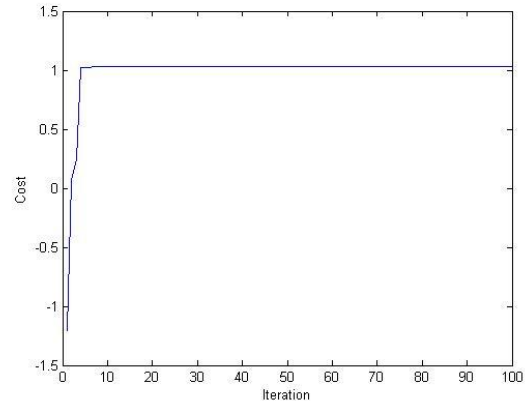

## Conclusion and Future Works

In this work, a framework for a new heuristic algorithm has been proposed. In this algorithm, the football system of European countries has been imitated. Three sorts of teams are considered in term of their financial strength: Wealthiest (strongest), Regular and Poorest (weakest). The strongest teams seek to buy the best players of regular teams and train them. If they work efficiently, they will remain, and otherwise, they will be sold and substituted with other players from regular teams in their special posts. In regular teams players are either bought or discovered, and for one year they will be trained. If their performances are spectacular, they will be sold to strongest teams or remained, and otherwise they will be substituted with a new player from mentioned sources. And finally in weakest teams, perfect players who show perfect outcome will be sold to other teams and otherwise, they will be substituted with other discovered young players. The algorithm, based on mentioned descriptions, show excellent performance in finding global optimums of the benchmarks. Its performance is completely competitive or better than other heuristics like GA and PSO. However, there are several points to investigate for further researches such as the impact of change I number of any sort of teams on its performance. Also there are many possible changes in transfers policies as now the algorithm works in the way that there is no difference between the good players of weakest teams to be sold. This point, in addition to the other possible cases, is the subjects of future researches.